\documentclass{article} 
\usepackage{nips15submit_e,times}
\usepackage{amsmath} 
\usepackage{bm}
\usepackage[top=1in, bottom=1in, left=1.25in, right=1.25in]{geometry}
\usepackage{cite,bbm}
\usepackage{url}
\usepackage{subfigure}
\usepackage{graphicx}
\usepackage{amssymb}
\usepackage{caption}
\usepackage{multirow}
\usepackage{authblk}
\usepackage{mathtools}
\usepackage{multirow}
\usepackage{fixmath}
\title{A C++ library for Multimodal Deep Learning}
\author[*]{Jian Jin\\Department of Computer Science\\ Johns Hopkins University}

\nipsfinalcopy 

\begin{document}
\maketitle
\thispagestyle{empty} 

\section{Multimodal Deep Learning Library}
MDL, Multimodal Deep Learning Library, is a deep learning framework that supports multiple models, and this document explains its philosophy and functionality. MDL runs on Linux, Mac, and Unix platforms. It depends on OpenCV.
\subsection{Modules}
MDL supports the following deep learning models: 
\begin{itemize}
\item Restricted Boltzmann Machine
\item Deep Neural Network
\item Deep Belief Network
\item Denoising Autoencoder
\item Deep Boltzmann Machine
\item Bimodal Model
\end{itemize}
It also provides interfaces for reading MNIST, CIFAR, and AVLetters data sets. Multiple options in learning and visualization interfaces are also provided.
\subsection{Document Content}
Chapter 2 introduces three graphical models. Chapters 3 to 7 give descriptions of each deep learning model. Chapter 8 introduces multimodal deep learning. Chapter 9 describes library architecture. Chapter 10 shows testing results.
\section{Network Survey}
This chapter includes a brief description of three graphical models that are mentioned in this document and a summary of deep learning models of MDL.
\subsection{Neural Network}
The Neural Network is a directed graph that consists of multiple layers of neurons, which is also known as units. Each pair of adjacent layers is connected, but there is no connection within each layer. The first layer is the input, called the visible layer $v$. Above the visible layer are multiple hidden layers $\{h_1, h_2,..., h_n\}$. The output of the last hidden layer forms the output layer $o$.\\
\\
In hidden layers, neurons in layer $h_{i}$ receive input from the previous layer,  $h_{i - 1}$ or $v$, and the output of $h_i$ is the input to the next layer, $h_{i+1}$ or $o$. The value transmitted between layers is called activation or emission. The activation is computed as:\begin{gather}
a^{(k)}_i = f((\sum_{j=1}^{n_{k-1}}a^{(k-1)}_jw^{k}_{ij})+b^{(k)}_i)=f(z^{(k)}_i)
\end{gather}
where $f$ is the activation function that defines a non-linear form of the output. A typical activation function is the sigmoid function. $z^{(k)}_i$ is the total input of the unit $i$ in layer $k$. It is computed as the weighted sum of the activations of the previous layer. The weight $w^{k}_{ij}$ and the bias $b^k_i$ are specified by the model.\\
\\
In forward propagation, activations are computed layer by layer. Training of the neural network uses backpropagation, which is a supervised learning algorithm. It calculates the error based on the network output and the training label, then uses this error to compute the gradients of error with respect to the weights and biases. The weights and biases are then updated by gradient descent.
\subsection{Markov Random Field}
The Markov Random Field, also called the Markov Network, is an undirected graphical model in which each node is independent of the other nodes given all the nodes connected to it. It defines the distribution of the variables in the graph.\\
\\
The Markov Random Field uses energy to decribe the distribution over the graph\begin{equation}
P(u) = \frac{1}{Z} e^{-E(u)},
\end{equation}
where $Z$ is a partition function defined by \begin{equation}
Z = \sum_{u} e^{-E(u)}.
\end{equation}
$E$ is the energy specified by the model, and $u$ is the set of states of all the nodes:\begin{equation}
u=\{v_1, v_2,...,v_n \}
\end{equation}
where $v_i$ is the state of node $i$.
\subsection{Belief Network}
The Belief Network, which is also called the Bayesian Network, is a directed acyclic graph for probabilistic reasoning. It defines the conditional dependencies of the model by associating each node $X$ with a conditional probability $P(X|Pa(X))$, where $Pa(X)$ denotes the parents of $X$. Here are two of its conditional independence properties:\\
\\
1. Each node is conditionally independent of its non-descendants given its parents.\\
2. Each node is conditionally independent of all other nodes given its Markov blanket, which consists of its parents, children, and children's parents.\\
\\
The inference of Belief Network is to compute the posterior probability distribution\begin{equation}
P(H|V) = \frac{P(H, V)}{\sum_H P(H,V)},
\end{equation}
where $H$ is the set of the query variables, and $V$ is the set of the evidence variables. Approximate inference involves sampling to compute posteriors. The Sigmoid Belief Network \cite{saul1996mean} is a type of the Belief Network such that \begin{equation}
P(X_i=1|Pa(X_i)) = \sigma(\sum_{X_j\in Pa(X_i)} W_{ji} X_j + b_i)
\end{equation}
where $W_{ji}$ is the weight assigned to the edge from $X_j$ to $X_i$, and $\sigma$ is the sigmoid function.
\subsection{Deep Learning Models}
There are many models in deep learning \cite{deeptutorial}. Below are included in MDL.\\
\\
The Restricted Boltzmann Machine is a type of Markov Random Field and is trained in an unsupervised fashion. It is the building block for other models and could be used for classification by adding a classifier on top of it.\\
\\
The Deep Neural Network is a neural network with multiple layers. Each layer is initialized by pretraing a Restricted Boltzmann Machine. Then fine tuning would refine the parameters of the model.\\
\\
The Deep Belief Network is a hybrid of the Restricted Boltzmann Machine and the Sigmoid Belief Network. It is a generative model but not a feedforward neural network nor a multilayer perceptron even though its training is similar to the Deep Neural Network.\\
\\
The Denoising Autoencoder is a type of neural network that has a symmetric structure. It could reconstruct the input data and if properly trained could denoise the input data. However, unlike neural networks, it could be trained with an unsupervised method.\\
\\
The Deep Boltzmann Machine is another type of Markov Random Field. It is pretrained by stacking Restricted Boltzmann Machines with adjusted weights and biases as an approximation to undirected graphs. Its fine tuning uses a method called mean field inference.
\section{Restricted Boltzmann Machine}
\subsection{Logic of the Restricted Boltzmann Machine}
\begin{figure}[h]
\centering
\includegraphics[height=1.5in]{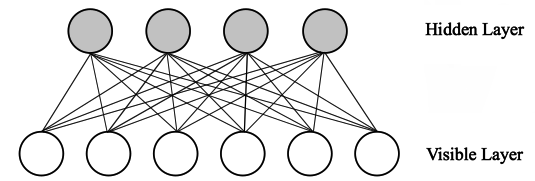}
\caption{Restricted Boltzmann Machine} \label{fig:side:a}
\end{figure}
The Restricted Boltzmann Machine (RBM) \cite{larochelle2008classification, le2008representational, salakhutdinov2007restricted} is a Markov Random Field consisting of one hidden layer and one visible layer. It is an undirected bipartite graph in which connections are between the hidden layer and the visible layer. Each unit $x$ is a stochastic binary unit such that\begin{equation}
state(x)=\begin{cases}1,~~~~p\\
0,~~~~1-p
\end{cases}
\end{equation}
where probability $p$ is defined by the model. Figure 3.1 shows a RBM with four hidden units and six visible units.\\
\\
As a Markov Random Field, a Restricted Boltzmann Machine defines the  distribution over the visible layer $v$ and the hidden layer $h$ as \begin{equation}
P(v,h) = \frac{1}{Z} e^{-E(v,h)},\label{eq:3}
\end{equation}
where $Z$ is the partition function defined as \begin{equation}
Z = \sum_{v, h} e^{-E(v,h)}.
\end{equation}
Its energy $E(v,h)$ is defined by \begin{equation}\label{eq:1}
E(v,h) = -\sum_i b^v_i v_i - \sum_j b^h_j h_j -\sum_i \sum_j v_i w_{i,j} h_j,
\end{equation}
where $b^v_i$ is the bias of the \emph{i}th visible unit and $b^h_j$ is the bias of the \emph{j}th hidden unit.\\
\\
Given the conditional independence property of the Markov Random Field, the distribution of one layer given the other layer could be factorized as
\begin{gather}
P(h|v)=\prod_j P(h_j|v) \label{eq:2}\\
P(v|h)=\prod_i P(v_i|h).
\end{gather}
Plug equation \eqref{eq:1} into equation \eqref{eq:3}: \begin{equation} \label{eq:5}
P(h|v)=\frac{P(h,v)}{\sum_{h'} P(h',v)}=\frac{exp(\sum_i b^v_i v_i + \sum_j b^h_j h_j +\sum_i \sum_j v_i w_{i,j} h_j)}{\sum_{h'} exp(\sum_i b^v_i v_i + \sum_j b^{h'}_j h'_j +\sum_i \sum_j v_i w_{i,j} h'_j)}.
\end{equation}
It is easy to show that \begin{equation}\label{eq:4}
\frac{exp(\sum_i b^v_i v_i + \sum_j b^h_j h_j +\sum_i \sum_j v_i w_{i,j} h_j)}{\sum_{h'} exp(\sum_i b^v_i v_i + \sum_j b^{h'}_j h'_j +\sum_i \sum_j v_i w_{i,j} h'_j)}=\prod_{j}\frac{exp(b^h_j + \sum_{i=1}^m W_{i,j} v_i )}{1+exp(b^h_j + \sum_{i=1}^m w_{i,j} v_i )}.
\end{equation}
By combining equation \eqref{eq:2}, \eqref{eq:5}, and \eqref{eq:4}, we have
\begin{equation}
P(h_j=1|v) = \sigma \left(b^h_j + \sum_{i=1}^{N_v} W_{i,j} v_i \right),
\end{equation}
where $\sigma$ is a sigmoid function\begin{equation}
\sigma(t) = \frac{1} {1 + exp(-t)}.
\end{equation}
Similarly,
\begin{equation}
P(v_i=1|h) = \sigma \left(b^v_i + \sum_{j=1}^{N_h} W_{i,j} h_j \right).
\end{equation}
\\
A Restricted Boltzmann Machine maximizes the likelihood $P(x)$ of the input data $x$, which is\begin{align}
P(x) = \sum_h P(x,h) = \sum_h \frac{1}{Z}e^{-E(h, x)}=\frac{1}{Z}e^{-F(x)},
\end{align}
where $F(x)$ is called Free Energy such that \begin{equation}
F(x) = - \sum_{i=1}^{n_v} b^v_ix_i-\sum_{j=1}^{n_h}\log(1+ exp(b^h_j+\sum_{k=1}^{n_v} W_{kj}x_k)).
\end{equation}
\\
In training, maximizing the likelihood of the training data is achieved by minimizing the negative log-likelihood of the training data. Because the direct solution is intractable, gradient descent is used, in which the weights $\{W_{ij}\}$, the biases of the hidden units $\{b^h_j\}$, and the biases of the visible units $\{b^v_i\}$ are updated. The gradient is approximated by an algorithm called Contrastive Divergence.\\
\\
The activation of the hidden layer is
\begin{equation}
a^h =\sigma(v * W + b^h),
\end{equation}
where $\sigma$ is performed in an element-wise fashion. The set of the hidden states, $\{h_j\}$, is expressed as a row vector $h$, the set of the hidden biases, $\{b^h_{j}\}$, is expressed as a row vector $b^ h$, and the set of the weights, $\{W_{ij}\}$, is expressed as a matrix $W$ which represents the weight from the visible layer to the hidden layer.\\
\\
The Restricted Boltzmann Machine has tied weights such that
\begin{equation}
a^v = \sigma(h * W^{T} + b^v).
\end{equation}
In training, the state of the visible layer is initialized as the training data.
\subsection{Training of the Restricted Boltzmann Machine}
In the training of Restricted Boltzmann Machine, the weights and the biases of the hidden and the visible layers are updated by gradient descent. Instead of stochastic gradient descent, in which each update is based on each data sample, batch learning is used. In batch learning, each update is based on a batch of training data. There are several epochs in training. Each epoch goes through the training data once.\\
\\
For instance, if the input data has 10,000 samples and the number of batches is 200, then there will be 200 updates in each epoch.  For each update, gradients will be based on 50 samples. If the number of epochs is 10, commonly there should be a total of 2000 updates in the training process. If the gradients computed are trivial, this process may stop earlier.\\
\\
The gradients of weights are given by Contrastive Divergence as: \begin{equation}\label{eq:7}
\nabla W_{ij} = \langle v_i * h_j\rangle_{recon} - \langle v_i * h_j\rangle_{data},
\end{equation}
where the angle brackets are expectations under the distribution specified by the
subscript. The expectations here are estimated by data sample mean, therefore
\begin{equation}
\nabla W_{ij} = \sum_{k = 1} ^ {m}((v_i * h_j)_{recon_k} - (v_i * h_j)_{data_k}) / m,
\end{equation}
where $m$ is the size of each data batch.\\
\\
States of the visible layer and the hidden layer form a sample in Gibbs sampling, in which the first sample gives states with the subscript ``data" in equation \eqref{eq:7}  and the second sample gives states with the subscript ``recon" in equation \eqref{eq:7}. Contrastive Divergence states that one step of Gibbs sampling, which computes the first and the second sample, approximates the descent with high accuracy. In RBM, Gibbs sampling works in the following manner:\\
\\
In Gibbs sampling, each sample $X = (x_1, \dots, x_n)$ is made from a joint distribution $\left.p(x_1, \dots, x_n)\right.$ by sampling each component variable from its posterior. Specifically, in the (\emph{i} + 1)th sample $X^{(i+1)} = (x_1^{(i+1)}, \dots, x_n^{(i+1)})$, $x_j^{(i+1)}$ is sampled from \begin{equation}
p(X_j|x_1^{(i+1)},\dots,x_{j-1}^{(i+1)},x_{j+1}^{(i)},\dots,x_n^{(i)}),
\end{equation}
in which the latest sampled component variables are used to compute posterior. Sampling each component variable $x_j$ once forms a sample.\\
\\
Each unit of RBM is a stochastic binary unit and its state is either 0 or 1. To sample $h_j$ from \begin{equation}
P(h_j=1|v) = \sigma \left(b_j + \sum_{i=1}^m W_{i,j} v_i \right),
\end{equation}
we can simply compute $a_j = \sigma \left(b_j + \sum_{i=1}^m W_{i,j} v_i \right)$. If $a_j$ is larger than a random sample from standard uniform distribution, state of $h_j$ is 1, otherwise 0. This method works because the probability that a random sample $u$ from standard uniform distribution is smaller than $a_j$  is $a_j$: \begin{equation}
P(u < a_j) = F_{\text{uniform}}(a_j) = a_j.
\end{equation}
So we have \begin{equation}
P(h_j=1|v) = P(u < a_j).
\end{equation}
Thus we could sample by testing if $u < a_j$, since it has probability $a_j$. Each unit has two possible states. If $u < a_j$ fails, the state is 0.\\
\\
In training, we first use the training data to compute the hidden layer posterior with \begin{equation}
P(h_j=1|v) = \sigma \left(b_j + \sum_{i=1}^m W_{i,j} v_i \right).
\end{equation}
The hidden layer states and the training data form the first sample. Then Gibbs sampling is used to compute the second sample.\\
\\
The gradient of the visible bias is \begin{equation}
\nabla b^v_{i} = \langle v_i\rangle_{recon} - \langle v_i\rangle_{data},
\end{equation}
and the gradient of the hidden bias is \begin{equation}
\nabla b^h_{j} = \langle h_j\rangle_{recon} - \langle h_j\rangle_{data}
\end{equation}
\subsection{Tricks in the Training}
\textbf{Dropout}\\
Dropout \cite{hinton2010practical} is a method to prevent neural networks from overfitting by randomly blocking emissions from a portion of neurons. It is similar to adding noise. In RBM training, a mask is generated and put on the hidden layer.\\
\\
For instance, suppose the hidden states are \begin{equation}
h = \{h_1, h_2, ..., h_n\}
\end{equation}
and the dropout rate is $r$. Then a mask $m$ is generated by 
\begin{eqnarray}
m_i=\begin{cases}
1,~~~~u_i > r\\
0,~~~~u_i \leq r
\end{cases}
\end{eqnarray}
where $u_i$ is a sample from standard uniform distribution, and $i\in \{1, 2, ..., n\}$. The emission of hidden layer would be \begin{equation}
\tilde{h} = h.* m
\end{equation}
where $.*$ denotes element-wise multiplication of two vectors. $\tilde{h}$, instead of $h$, is used to calculate visible states.\\
\\
\textbf{Learning Rate Annealing}\\
There are multiple methods to adapt the learning rate in gradient descent. If the learning rate is trivial, updates may tend to get stuck in local minima and waste computation. If it is too large, the updates may bound around minima and could not go deeper. In annealing, learning rate decay adjusts the learning rate for better updates.\\
\\
Suppose $\alpha$ is the learning rate. In exponential decay, the annealed rate is\begin{equation}
\alpha_a = \alpha * e^{-kt},
\end{equation}
where $t$ is the index of the current epoch, $k$ is a customized coefficient. In divide decay, the annealed rate is\begin{equation}
\alpha_a = \alpha / (1 + kt).
\end{equation}
A more common method is step decay:\begin{equation}
\alpha_a = \alpha * 0.5^{\lfloor{t / 5}\rfloor{}}.
\end{equation}
where learning rate is reduced by half every five epochs. The coefficients in the decay method should be tuned in testings.\\
\\
\textbf{Momentum}\\
With momentum $\rho$, the update step is \begin{equation}
\Delta_{t+1} W = \rho * \Delta_t W - r\nabla W.
\end{equation}
The update value is a portion of previous update value minus the gradient. Thus, if the gradient is in the same direction of the previous update, the update value will become larger. If the gradient is in the different direction, the variance of the updates will decrease. In this way, time to converge is reduced.\\
\\
Momentum is often applied with annealing so that steps of updates will not be too large. A feasible scheme is \begin{eqnarray}
\rho=\begin{cases}
0.5,~~~~t <  5\\
0.9,~~~~t \geq 5
\end{cases}
\end{eqnarray}
where $t$ is the index of the current epoch.\\
\\
\textbf{Weight Decay}\\
In weight decay, a penalty term is added to the gradient as a regularizer. $L_1$ penalty $p_1$ is \begin{equation}
p_1 = k \times \sum_{i,j} |W_{ij}|.
\end{equation}
$L_1$ penalty causes many weights to become zero and a few weights to become large. $L_2$ penalty $p_2$ is \begin{equation}
p_2 = k \times \sum_{i,j} W_{ij}^2.
\end{equation}
$L_2$ penalty causes weights to become more even and smaller. The coefficient $k$ is customized, sometimes 0.5 would work. Penalty must be, as well as the gradient, multiplied by the learning rate so that annealing will not change the model trained. 
\subsection{Classifier of the Restricted Boltzmann Machine}
A classifier based on RBM could be built by training a classifier layer with the softmax function on top of the hidden layer. The softmax function takes a vector $a = \{a_1, a_2,..., a_q\}$ as input, and outputs a vector $c = \{c_1, c_2,..., c_q\}$ with the same dimension, where\begin{equation}
c_i = \frac{e^{a_i}}{\sum_{k=1}^q e^{a_k}}.
\end{equation}
\\
One-of-$K$ scheme is used to present the class distribution. If there are $K$ classes in the training data, then the label of a sample in class $i$ is expressed as a vector of length $K$ with only the $i$th element as 1, the others as 0. For instance, if the training data has five classes, a sample in the fourth class has the label\begin{equation*}
\{0,0,0,1,0\}.
\end{equation*}
\\
For training data in $K$ classes, there are $K$ neurons in the classifier layer. For each data sample, the softmax activation emits a vector of length $K$. The index of the maximum element in this emission is the label. The elements of the softmax activation sum to 1 and the activation is the prediction distribution:\begin{equation}
c_i = P\{\text{The sample is in class i}\}.
\end{equation}
\\
The input of the softmax function is of $K$-dimension whereas there may be hundreds of units in the hidden layer. So the projection from the hidden layer to the classifier should be learned. If the hidden layer has $n_h$ units and the training data is in $K$ classes, the weight of this projection should be a matrix of dimension $n_h\times K$.\\
\\
Backpropagation with the combination of cross entropy loss and softmax function is used to update this weight. Using $t$ to represent the labels and $c$ to represent the prediction distribution, the cross entropy loss is \begin{equation}
L = -\sum_{i = 0}^K t_i log(c_i).
\end{equation}
\\
By chain rule:\begin{equation}
\frac{\partial L}{\partial W_{ij}} = \sum_{p=1}^K \frac{\partial L}{\partial c_p} \frac{\partial c_p}{\partial W_{ij}}
\end{equation}
and \begin{equation}
\frac{\partial c_p}{\partial W_{ij}}= \sum_{q=1}^K \frac{\partial c_p}{\partial z_q} \frac{\partial z_q}{\partial W_{ij}}
\end{equation}
where $c$ is the output of softmax activation and $z$ is its input. That is to say, \begin{equation}
c_i = \frac{exp(z_i)}{\sum_{p=1}^K exp(z_p)}.
\end{equation} Furthermore\begin{equation}
\frac{\partial L}{\partial c_p} = -\frac{t_p}{c_p}
\end{equation}
\begin{equation}
\frac{\partial c_p}{\partial z_q} = \begin{cases}
c_p(1-c_p)~~~~~~\text{if } p = q
\\
-c_p\times c_q~~~~~~~~\text{if } p \neq q
\end{cases}
\end{equation}
\begin{equation}
\frac{\partial z_q}{\partial W_{ij}} = \begin{cases}
a^h_i~~~~~~\text{if } q = j
\\
0~~~~~~~~\text{if } q \neq j
\end{cases}
\end{equation}
where $a^h_i$ is the activation of the hidden layer. Based on the above equations, for the combination of the cross entropy loss and the softmax activation \begin{equation}
\frac{\partial L}{\partial W_{ij}} = a_i^h(c_j - t_j).
\end{equation}
Expressed in row vectors it is \begin{equation}
\nabla W = (a^h)^T\times (c - t),
\end{equation}
where $(a^h)^T$ is the transpose of row vector $a^h$. This gradient is used in batch learning.
\subsection{Implementation of the Restricted Boltzmann Machine}
Class rbm is implemented in the header file rbm.hpp. Here is a selected list of its methods:\\
\\
I. void \textbf{dropout}(double i)\\
-Set dropout rate as input i.\\
\\
II. void \textbf{doubleTrain}(dataInBatch $\&$trainingSet, int numEpoch, int inLayer, ActivationType at = sigmoid$\_$t, LossType lt = MSE, GDType gd = SGD, int numGibbs = 1)\\
-Train an RBM layer in Deep Boltzmann Machine, which is an undirected graph. This part will be explained in DBM section.\\
\\
III. void \textbf{singleTrainBinary}(dataInBatch $\&$trainingSet, int numEpoch, ActivationType at = sigmoid$\_$t, LossType lt = MSE, GDType gd = SGD, int numGibbs = 1);\\
-Train an RBM layer with binary units in Deep Belief Networks and RBM.\\
\\
IV. void \textbf{singleTrainLinear}(dataInBatch $\&$trainingSet, int numEpoch, ActivationType at= sigmoid$\_$t, LossType lt = MSE, GDType gd = SGD, int numGibbs = 1);\\
-Train an RBM layer with linear units.\\
\\
V. void \textbf{singleClassifie}r(dataInBatch $\&$modelOut, dataInBatch $\&$labelSet, int numEpoch, GDType gd = SGD);\\
-Build a classifier layer for RBM.\\
\\
The model could be tested by running runRBM.cpp. First train a RBM with one hidden layer:\\
\\
RBM rbm(784, 500, 0);\\
rbm.dropout(0.2);\\
rbm.singleTrainBinary(trainingData, 6);\\
dataInBatch modelOut = rbm.g$\_$activation(trainingData);\\
\\
The hidden layer has 500 units, and the index of the RBM is 0. The dropout rate is 0.2. After training, stack the classifier layer with the softmax function on top of it:\\
\\
RBM classifier(500, 10, 0);\\
classifier.singleClassifier(modelOut, trainingLabel, 6);\\
classificationError e = classifyRBM(rbm, classifier, testingData, testingLabel, sigmoid$\_$t);\\
\\
MNIST dataset has ten classes, so there are ten units in the classifier layer.
\subsection{Summary}
RBM is the foundation for several multi-layer models. It is crucial that this component is correctly implemented and fully understood. The classifier may be trained with the hidden layer at the same time. Separating these two processes facilitates checking problems in the implementation.

\section{Deep Neural Network}
\subsection{Pretraining of the Deep Neural Network}
\begin{figure}[h]
\centering
\includegraphics[height=3.2in]{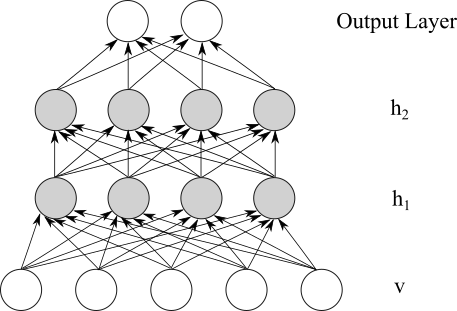}
\caption{Deep Neural Network} \label{fig:side:a}
\end{figure}
A Deep Neural Network (DNN) \cite{becker1992self} is a neural network with multiple hidden layers. In neural networks, initialization of weights could significantly affect the training results. Pretraining, in which multiple Restricted Boltzmann Machines are trained to initialize parameters of each DNN layer, provides weight initialization that saves training time. Backpropagation is a time-consuming process. With pretraining, the time consumed by backpropagation could be significantly reduced. Figure 4.1 shows a DNN with two hidden layers.\\
\\
Below shows steps to build a Deep Neural Network for classification:\\
\\
I. Set the architecture of the model, specifically the size of the visible layer and the hidden layers, \(n_0, n_1, n_2, ..., n_N\). $n_0$ is the input dimension and input forms a visible layer. $n_N$ equals to the number of classes in training data.\\
\\
II. Pretrain hidden layers:\\
for $i = 1\text{ to }N $:\\
\indent 1. Train an RBM with the following settings:\\
\begin{gather*}
n_h^{(i)}=n_i\\
n_v^{(i)}=n_{i-1}\\
d_i=a_{i-1}\\
\end{gather*}
\indent where
\begin{gather*}
n_h^{(i)} = \text{Dimension of hidden layer of RBM trained for layer }i,\\
n_v^{(i)} = \text{Dimension of visible layer of RBM trained for layer }i,\\
d_i = \text{Input of RBM trained for layer i},\\
a_{i-1} = \text{Activation of the }(i-1)\text{th DNN layer}.
\end{gather*}
\indent 2. Set \begin{gather*}
W_i = W_{RBM},\\
b_i = b^h_{RBM},\\
a_i = a_{RBM}.
\end{gather*}
\indent RBM is used to initialize weights and biases of each DNN layer.\\
end for.\\
\\
III. Fine Tuning:\\
Use backpropagation to refine the weights and the biases of each layer. In backpropagation, one epoch goes through training data once. A dozen of epochs may suffice.\\
\\
Classification with Deep Neural Network is similar to RBM. The last layer, which uses softmax activation, gives the prediction distribution.
\subsection{Fine Tuning of the Deep Neural Network}
Fine tuning uses backpropagation:\\
\\
I. Perform a pass through all the layers. Compute total inputs of each layer $\{z^{(1)},..., z^{(N)}\}$ and activations of each layer $\{a^{(1)},..., a^{(N)}\}$. $a^{(i)}$ is the row vector that represents the activation of layer $i$. \\
\\
II. For the last layer, compute $\delta_i^{(N)}$ as
\begin{equation}
\delta_i^{(N)} = \frac{\partial L}{\partial z^{(N)}_i}
\end{equation}
where $L$ is the classification error. A row vector $\delta^{(N)}$ is computed. \\
\\
III. For $l$ = $N-1, ...,1$, compute 
\begin{align}
\delta^{(l)} = \left(\delta ~ (W^{(l)})^T \right) \bullet g'(z^{(l)})
\end{align}
where $g$ is the activation function.\\
\\
IV. Compute the gradients in each layer. For $l$ = $N, ..., 1$, compute 
\begin{align}
\nabla_{W^{(l)}} L &= (a^{(l-1)})^T~\delta^{(l)}, \\
\nabla_{b^{(l)}} L &= \delta^{(l)}.
\end{align}
where $a^{(0)}$ is the training data.\\
\\
V. Update the weights and biases of each layer using gradient descent.\\
\\
In fine tuning, the training data is repeatedly used to refine the model parameters.
\subsection{Implementation of Deep Neural Network}
Class dnn is implemented in the header file dnn.hpp. Class RBMlayer provides interfaces to store architecture information. Here is a selected list of methods of class dnn:\\
\\
I. void \textbf{addLayer}(RBMlayer $\&$l)\\
Add a layer to the current model. This object of class RBMlayer should store information of layer size, weight, bias, etc.\\
\\
II. void \textbf{setLayer}(std::vector$<$size$\_$t$>$ rbmSize)\\
Object of class dnn could automatically initialize random weights and biases of each layer by inputting a vector of layer sizes.\\
\\
III. void \textbf{train}(dataInBatch $\&$trainingData, size$\_$t rbmEpoch, LossType l = MSE, ActivationType a = sigmoid$\_$t)\\
This method trains all the layers without classifier. The structure of dnn should be initialized before calling this method. \\
\\
IV. void \textbf{classifier}(dataInBatch $\&$trainingData, dataInBatch $\&$trainingLabel, size$\_$t rbmEpoch, int preTrainOpt, LossType l = MSE, ActivationType a = sigmoid$\_$t)\\
Build a Deep Neural Network with a classifier layer. This function contains pretraining option preTrainOpt. If preTrainOpt=1, pretrain each layer by training RBMs, else randomly initialize layer parameters without pretraining.\\
\\
V. void \textbf{fineTuning}(dataInBatch $\&$label, dataInBatch $\&$inputSet, LossType l)\\
Fine tuning step that uses backpropagation.\\
\\
VI. classificationError \textbf{classify}(dataInBatch $\&$testingSet, dataInBatch $\&$testinglabel);\\
Perform Classification. The result is stored in the format classficationError.
\subsection{Summary}
The training of the Deep Neural Network consists of two steps: pretraining by stacking multiple RBMs, and fine tuning with backpropagation. Pretraining significantly reduces time spent on backpropagation.

\section{Deep Belief Network}
\subsection{Logic of the Deep Belief Network}
\begin{figure}[h]
\centering
\includegraphics[height=2.8in]{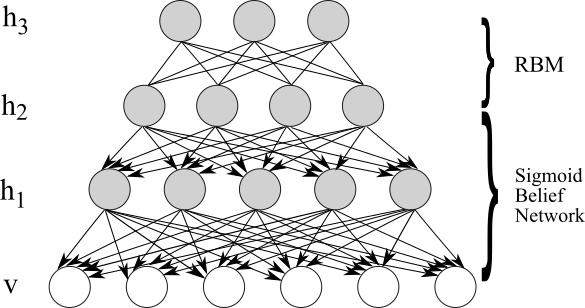}
\caption{Deep Belief Network} \label{fig:side:a}
\end{figure}
A Deep Belief Network (DBN) is a hybrid of a Restricted Boltzmann Machine and a Sigmoid Belief Network. A Deep Belief Network maximizes the likelihood $P(x)$ of the input $x$. Figure 5.1 shows a DBN.\\
\\
For a Deep Belief Network with $N$ hidden layers, the distribution over the visible layer (input data) and hidden layers is\begin{equation}
P(v, h_1, ..., h_N)  = P(v| h_1) \times\left(\prod_{k=1}^{N-2}  P(h_k| h_{k+1}))\right)\times P(h_{N-1},h_N).
\end{equation}
To prove this, by chain rule we can write\begin{equation}
P(v, h_1, ..., h_N)  = P(v| h_1, ..., h_N) \times\left(\prod_{k=1}^{N-2}  P(h_k| h_{k+1},..., h_{N}))\right)\times P(h_{N-1},h_N).
\end{equation}
Additionally, each node is independent of its ancestors given its parent. So these hold:\begin{gather}
P(v| h_1, ..., h_N) = P(v|h_1),\\
P(h_k| h_{k+1},..., h_{N}) = P(h_k| h_{k+1}).
\end{gather}
Then we have \begin{equation}
P(v, h_1, ..., h_N)  = P(v| h_1) \times\left(\prod_{k=1}^{N-2}  P(h_k| h_{k+1}))\right)\times P(h_{N-1},h_N),
\end{equation}
where $P(v| h_1) \times\left(\prod_{k=1}^{N-2}  P(h_k| h_{k+1}))\right)$ is the distribution over the Sigmoid Belief Network and $P(h_{N-1},h_N)$ is the distribution over Restricted Boltzmann Machine.\\
\\
For classification there should be a layer $y$ on top of the last hidden layer. With  layer $y$ that represents prediction distribution, the distribution over the Deep Belief Network is
\begin{equation}
P(v, h_1, ..., h_N, y)  = P(v| h_1) \times\left(\prod_{k=1}^{N-2}  P(h_k| h_{k+1}))\right)\times P(h_{N-1},h_N, y).
\end{equation}
where $P(h_{N-1},h_N, y)$ is the distribution over a RBM which has labels $y$ and the state $h_{N-1}$ as the input. In the pretraining of the Deep Belief Network, RBMs are stacked. In fine tuning, Up-Down algorithm is used.
\subsection{Training of the Deep Belief Network}
In training, the Deep Belief Network should maximize the likelihood of the training data. With the concavity of the logarithm function, the lower bound of the log likelihood of the training data $x$ could be found:\begin{equation}\label{eq:10}
\log P(x)=\log\left(\sum_h Q(h|x)\frac{P(x,h)}{Q(h|x)}\right)\geq \sum_h Q(h|x)\log\frac{P(x,h)}{Q(h|x)},
\end{equation}
and
\begin{equation}\label{eq:8}
\sum_h Q(h|x)\log\frac{P(x,h)}{Q(h|x)}= \sum_h Q(h|x)\log P(x, h)-\sum_h Q(h|x)\log Q(h|x),
\end{equation}
where $Q(h|x)$ is an approximation to the true probability $P(h|x)$ of the model.\\
\\
If $Q(h|x) = P(h|x)$, the right-hand side of \eqref{eq:8} could be written as \begin{equation}\label{eq:11}
\begin{split}
&\sum_h P(h|x)(\log P(h|x) + \log P(x))-\sum_h P(h|x)\log P(h|x)\\
& = \sum_h P(h|x)\log P(x) =\log P(x)\sum_h P(h|x) =\log P(x).
\end{split}
\end{equation}
Combining equation \eqref{eq:11}, \eqref{eq:8} with \eqref{eq:10}, we could find that when $Q(h|x) = P(h|x)$, the lower bound is tight.\\
\\
Moreover, the more different $Q(h|x)$ is from $P(h|x)$, the less tight the bound is. The lower bound of \eqref{eq:10} could be expressed as \begin{equation}
\log P(x) - KL(Q(h|x)||P(h|x)).
\end{equation}
If the approximation $Q(h|x)$ becomes closer to the true posterior $P(h|x)$,  their KL divergence will be smaller, and the bound will be higher. Unlike the true posterior, the approximation could be factorized as \begin{equation}
Q(h|x)=\prod_{i=1}^{n_h} Q(h_i|x).
\end{equation}
\\
Consequently, in training the goal is to find approximation $Q(h|x)$ with high accuracy and at the same time maximizes the bound. This could be done by stacking pretrained RBMs. The lower bound of \eqref{eq:10} could be factorized as
\begin{equation}\label{eq:9}
\sum_h Q(h|x)\log\frac{P(x,h)}{Q(h|x)}= \sum_h Q(h|x)(\log P(x|h) + \log P(h))-\sum_h Q(h|x)\log Q(h|x)
\end{equation}
In the right-hand side of equation \eqref{eq:9}, $Q(h|x)$ and $P(x|h)$ are given by the first pretrained RBM. Therefore, to maximize the lower bound is to maximize \begin{equation}
\sum_h Q(h|x)\log P(h).
\end{equation}
Since the RBM maximizes the likelihood of the input data, staking another RBM on top of the first hidden layer maximizes the lower bound. Moreover, \begin{equation}
P(h) = \sum_{h^{(2)}}P(h, h^{(2)}), 
\end{equation}
where $h^{(2)}$ is computed by the second pretrained RBM. The second RBM takes sample made from $Q(h|x)$ as input. But it could be trained independently since its parameters do not depend on the parameters of the first pretrained RBM. This is why greedy layer-wise pretraining works.\\
\\
In fine tuning, Up-Down algorithm \cite{hinton2006fast} is used. It is a combination of Contrastive Divergence and Wake-Sleep algorithm \cite{hinton1995wake}. Wake-Sleep algorithm is used in the learning of the Sigmoid Belief Network.
\subsection{Classification of the Deep Belief Network}
In training of the Restricted Boltzmann Machine, dropout is used to alleviate overfitting. This method reminds us that RBM could predict missing values.\\
\\
In a deep belief network, each approximation $Q(h_{k+1}|h_k)$ could be computed based on states $h_k$ and model parameters. In a deep belief network with classifier, the top RBM takes labels and hidden layer $h_{N-1}$ to compute states of the last hidden layer $h_N$, which is illustrated in Figure 5.2.\\
\\
Suppose $l$ is the set of units that represents the prediction distribution. In classification, we can fill $l$ with zeros, compute approximation $Q(h_N|l, h_{N-1})$, and then use the states $h_N$ sampled from $Q(h_N|l, h_{N-1})$ to compute the prediction distribution $l$ with $P(l, h_{N-1}| h_N)$.
\begin{figure}[h]
\centering
\includegraphics[height=5.0in]{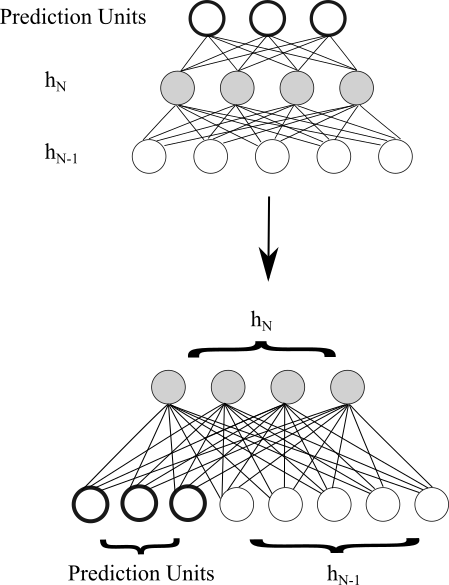}
\caption{Top RBM of DBN with classifier} \label{fig:side:a}
\end{figure}
\subsection{Implementation of the Deep Belief Network}
Class dbn is in the header file dbn.hpp. Here is a selected list of its methods:\\
\\
I. void \textbf{addLayer}(RBMlayer $\&$l)\\
Add a layer to the current model. This object of class RBMlayer should store information of layer size, weight, bias, etc. It could also be modified after added to  the model.\\
\\
II. void \textbf{setLayer}(std::vector$<$size$\_$t$>$ rbmSize)\\
Object of class dnn could automatically initialize random weights and biases of each layer by inputting a vector of layer sizes.\\
\\
III. void \textbf{train}(dataInBatch $\&$trainingData, size$\_$t rbmEpoch, LossType l = MSE, ActivationType a = sigmoid$\_$t)\\
This method trains a dbn without classifier. The architecture of dbn should be initialized before calling this method. \\
\\
IV. void \textbf{classifier}(dataInBatch $\&$trainingData, dataInBatch $\&$trainingLabel, size$\_$t rbmEpoch, LossType l = MSE, ActivationType a = sigmoid$\_$t)\\
This method trains a dbn with classifier. The architecture of dbn should be initialized before calling this method.\\
\\
V. void \textbf{fineTuning}(dataInBatch $\&$dataSet, dataInBatch $\&$labelSet, int epoch)\\
The fine tuning uses Up-Down algorithm.\\
\\
VI. classificationError \textbf{classify}(dataInBatch $\&$testingSet, dataInBatch $\&$testinglabel);\\
Perform Classification.
\subsection{Summary}
It is easy to confuse the Deep Belief Network with the Deep Neural Network. Both of them stack RBMs in the pretraining. However, because their principles and structures are distinct, their fine tunining and classification methods are different.

\section{Denoising Autoencoder}
\subsection{Training of the Autoencoder}
\begin{figure}[h]
\centering
\includegraphics[height=3.2in]{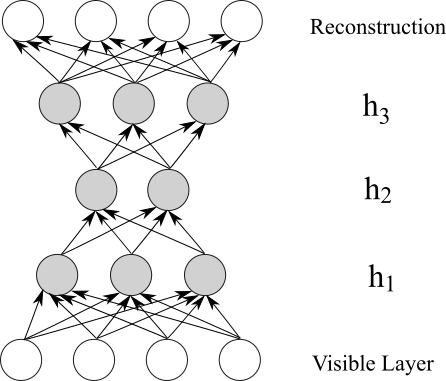}
\caption{Autoencoder} \label{fig:side:a}
\end{figure}
Autoencoder(AE) \cite{hinton2006reducing} is a type of neural network forming  a directed graph. Its symmetricity requires that in an autoencoder with $(N + 1)$ layers (including visible layer and output layer), the dimension of each layer is constrained by \begin{gather}
n_i = n_{N - i}~~~~\text{for }0\leq i\leq N
\end{gather}
where $n_i$ is the dimension of the $i$th layer, and $n_0$ is the input dimension. Since the output layer and the visible layer are of the same dimension, it is expected that autoencoders could reconstruct the input data. Thus, the training of autoencoders is unsupervised learning because input data is used as labels in fine tuning, and reconstruction errors could be used to access the model. Autoencoders could also be used to construct classifiers by adding a classifier layer on top of it. Figure 6.1 shows an Autoencoder.\\
\\
Below is how to construct an autoencoder:\\
I. Set the architecture of the model, specifically the size of each layer, \(n_0, n_1, n_2, ..., n_N\).\\
\\
II. Pretraining:\\
for $i = 1\text{ to }N/2$:\\
\indent 1. Train an RBM with the following settings:\\
\begin{gather*}
n_h^{(i)}=n_i\\
n_v^{(i)}=n_{i-1}\\
d_i=a_{i-1}
\end{gather*}
\indent where
\begin{gather*}
n_h^{(i)} = \text{Dimension of hidden layer of RBM trained for layer }i,\\
n_v^{(i)} = \text{Dimension of visible layer of RBM trained for layer }i,\\
d_i = \text{Input of RBM trained for layer i},\\
a_{i-1} = \text{Activations of the }(i-1)\text{th layer of autoencoder},
\end{gather*}
\indent 2. Initialize parameters of the current layer
\begin{gather*}
W_i = W_{RBM},\\
b_i = b^h_{RBM},\\
a_i = a_{RBM}.
\end{gather*}
\indent The parameters of trained RBM are used to initialize the parameters of the layer.\\
end for.\\
\\
for $i = N/2+1\text{ to }N$:\\
\indent Initialize parameters \begin{gather*}
W_i = W_{N - i}^{T},\\
b_i = b_{N-i}.
\end{gather*}
end for.\\
\\
III. Fine Tuning:\\
Backpropagation with Mean Squared Error. The error is computed based on the reconstruction and the training data.
\subsection{Fine tuning of the Autoencoder}
Fine tuning of Autoencoder uses backpropagation, which is:\\
\\
I. Perform a forward propagation through all layers that computes layer inputs $\{z^{(1)},..., z^{(N)}\}$ and activations $\{a^{(1)},..., a^{(N)}\}$. $a^{(i)}$ is a row vector representing the activation of layer $i$. \\
\\
II. For the last layer, compute $\delta_i^{(N)}$ as
\begin{equation}
\delta_i^{(N)} = \frac{\partial L}{\partial z^{(N)}_i}
\end{equation}
where $L$ is the reconstruction error. This step computes a row vector $\delta^{(N)}$. \\
\\
III. For $l$ = $N-1, ..., 1$, compute 
\begin{align}
\delta^{(l)} = \left(\delta^{(l+1)} ~ (W^{(l)})^T \right) \bullet g'(z^{(l)})
\end{align}
where $g$ is the activation function and here $g'$ is performed in an element-wise fashion.\\
\\
IV. Compute the gradients in each layer. For $l$ = $N, ..., 1$, compute 
\begin{align}
\nabla_{W^{(l)}} L &= (a^{(l-1)})^T~\delta^{(l)}, \\
\nabla_{b^{(l)}} L &= \delta^{(l)}.
\end{align}
where $a^{(0)}$ is the input data of the autoencoder.\\
\\
V. Update the weights and biases of each layer with gradient descent.\\
\\
The reconstruction error of Autoencodr is \begin{equation}
L = \frac{1}{2}\sum_{i=1}^{n_N}(a^{(N)}_i - a^{(0)}_i)^2
\end{equation}
where \begin{gather*}
n_{N} = \text{Dimension of the output/reconstruction},\\
a^{(N)}_{i} = \text{Activation of the }i\text{th unit in the output layer},\\
a^{(0)}_{i} = \text{Activation of the }i\text{th unit in the visible layer}.
\end{gather*}
Backpropagation involves computing $\{\delta^{(i)}\}$, which is based on the derivatives of activation function and error function. Here is how to compute these two values:\\
\\
For sigmoid activation:
\begin{equation}
g'(t) = \frac{\partial{(1+e^{-t})^{-1}}}{\partial{t}} = \frac{1}{1+e^{-t}}\frac{e^{-t}}{1+e^{-t}}.
\end{equation}
That is to say\begin{equation}
g'(z_i) = \frac{1}{1+e^{-z_i}}(1-\frac{1}{1+e^{-z_i}})=a_i(1-a_i).
\end{equation}
For $\delta_i^{(N)}$, by chain rule,\begin{equation}
\delta_i^{(N)} = \frac{\partial L}{\partial z^{(N)}_i}=\sum_{p=1}^{n_N} \frac{\partial L}{\partial a^{(N)}_p} \frac{\partial a^{(N)}_p}{\partial z^{(N)}_i} = \frac{\partial L}{\partial a^{(N)}_i} \frac{\partial a^{(N)}_i}{\partial z^{(N)}_i} = (a^{(N)}_i-a^{(0)}_i)\times a^{(N)}_i(1-a^{(N)}_i).
\end{equation}
\subsection{Denoising Autoencoder}
The Denoising Autoencoder reconstructs the corrupted data, and could predict missing values. By adding noises to corrupt the training data, we could transform the Autoencoders to the Denoising Autoencoders. Here is how to make this transformation:\\
\\
Firstly a denoise rate $r$ is chosen, and the mask is made as follows:\begin{equation}
m_i=\begin{cases} 
1,~~~~\text{if }U_i > r\\
0,~~~~\text{otherwise}\\
\end{cases}
~~1\leq i\leq n_N\\
\end{equation}
where $U_i$ is the $i$th sample from standard uniform distribution, $n_N$ is the size of the last layer, which is also the dimension of input data.\\
\\
Secondly compute the corrupted input data\begin{equation}
a^{(c)} = a^{(0)}\cdot m = \sum_{i=1}^{n_N} a^{(0)}_im_i
\end{equation}\\
\\
Finally use $a^{(c)}$ as the training data to compute the reconstruction and still use the uncorrupted data $a^{(0)}$ as labels in fine tuning. In reconstruction, fine tuning makes more improvements in the Denoising Autoencoder than in the Autoencoder, and is crucial to the Denoising Autoencoder.
\subsection{Implementation of the Denoising Autoencoder}
The implementation of DAE is in the header file autoencoder.hpp.\\
\\
I. void \textbf{addLayer}(RBMlayer $\&$l)\\
Add a layer to current DAE. This object of class RBMlayer should store information of layer size, weight, bias, etc. It could also be modified after added to DAE.\\
\\
II. void \textbf{setLayer}(std::vector$<$size$\_$t$>$ rbmSize)\\
Object of class AutoEncoder could automatically initialize random weights and biases of each layer by inputting a vector of layer sizes.\\
\\
III. void \textbf{train}(dataInBatch $\&$trainingData, size$\_$t rbmEpoch, LossType l = MSE, ActivationType a = sigmoid$\_$t)\\
This method trains all the layers without classifier. The structure of this AutoEncoder object should be initialized before calling this method. \\
\\
IV. void \textbf{reconstruct}(dataInBatch $\&$testingData)\\
This method gives the reconstruction of testingData. It should be called only after the model has been trained.\\
\\
V. dataInBatch \textbf{g$\_$reconstruction}()\\
Get the reconstruction.\\
\\
VI. void \textbf{fineTuning}(dataInBatch $\&$originSet, dataInBatch $\&$inputSet, LossType l)\\
Use backpropagation. Unlike DBN, argument LossType should be MSE instead of CrossEntropy.\\
\\
VI. void \textbf{denoise}(double dr)\\
Set the denoise rate as dr. When model is in training, it will detect if denoise rate is set. So if this method is called before training, the Denoising Autoencoder will be trained automatically. Otherwise the Autoencoder will be trained.
\subsection{Summary}
In pretraining, parameters of half of the layers are acquired by stacking RBMs. The parameters of the other half are given by the symmetric architecture. Fine tuning is crucial to the Denoising Autoencoder because it uses uncorrupted data to modify the model trained with corrupted data. The performance of the Denoising Autoencoder is straightforward to assess because one could observe the reconstructed images.

\section{Deep Boltzmann Machine}
\subsection{Logic of the Deep Boltzmann Machine}
\begin{figure}[h]
\centering
\includegraphics[height=2.8in]{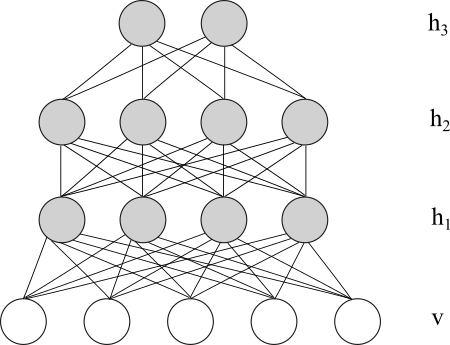}
\caption{Deep Boltzmann Machine} \label{fig:side:a}
\end{figure}
A Deep Boltzmann Machine(DBM) \cite{salakhutdinov2009deep} is a Markov Random Field with multiple layers. Connections exist only between adjacent layers. Intuitively, it could incorporate top-down feedback when computing bottom-up approximations. Figure 7.1 shows a DBM.\\
\\
The energy function of a Deep Boltzmann Machine with $N$ hidden layers is\begin{equation}
E(v, h^{(1)},..h^{(N)})=-v^{T}W^{(1)}h^{(1)} -(h^{(1)})^TW^{(2)}h^{(2)}-...- (h^{(N-1)})^TW^{(N)}h^{(N)}
\end{equation}
where $W^{(i)}$ is the weight from the previous layer to the $i$th hidden layer.\\
\\
A Deep Boltzmann Machine maximizes the likelihood of the input data. The gradient of its log likelihood is \begin{equation}
\frac{\partial \log P(v)}{\partial W^{(i)}} = \langle h^{(i-1)}(h^{(i)})^T\rangle_{data} - \langle h^{(i-1)}(h^{(i)})^T\rangle_{model}.
\end{equation}
\subsection{Pretraining of the Deep Boltzmann Machine}
Because the Deep Boltzmann Machine is an undirected model, the last hidden layer receives input from the previous adjacent layer, and the other hidden layers receive inputs from both directions. So when training Restricted Boltzmann Machines, the weights and biases need to be adjusted for better approximations. The pretraining process is as follows:\\
\\
I. Set the architecture of the model, specifically the size of each layer, \(n_0, n_1, n_2, ..., n_N\). $n_0$ is the dimension of the training data.\\
\\
II. Pretrain the first hidden layer:\\
Train an RBM, in which the weight from the visible layer $v$ to the hidden layer $h_1$ is $2W_1$ and the weight from $h_1$ to $v$ is $W_1^{T}$. $W_1$ is the weight of the first DBM hidden layer.\\
\\
III. Pretrain intermediate hidden layers:\\
for $i = 2\text{ to }N - 1$:\\
\indent 1. Train an RBM with the following settings:\\
\begin{gather*}
n_h^{(i)}=n_i\\
n_v^{(i)}=n_{i-1}\\
d_i=a_{i-1}\\
\end{gather*}
\indent where
\begin{gather*}
n_h^{(i)} = \text{Dimension of hidden layer of RBM trained for layer }i,\\
n_v^{(i)} = \text{Dimension of visible layer of RBM trained for layer }i,\\
d_i = \text{Input of RBM trained for layer i},\\
a_{i-1} = \text{Activation of the }(i-1)\text{th layer}.
\end{gather*}
\indent 2. Set \begin{gather*}
W_i = W_{RBM} / 2,\\
b_i = b^h_{RBM} / 2,\\
a_i = a_{RBM}.
\end{gather*}
\indent Weights and biases are adjusted here for better approximations.\\
\\
IV. Pretrain the last hidden layer:\\
Train an RBM, in which the weight from the hidden layer $h_{N-1}$ to the hidden layer $h_N$ is $W_N$ and the weight from $h_{N}$ to $h_{N-1}$ is $2W_N^{T}$. $W_N$ is the weight of the last hidden layer of DBM.
\subsection{Mean Field Inference}
The mean field inference \cite{salakhutdinov2010efficient
} of the Deep Boltzmann Machine involves iterative updates of the approximations $Q$. It is performed after pretraining. The algorithm is as follows:\\
\\
\noindent \textbf{Algorithm} Mean Field Inference\\
Initialize $M$ samples $\{\tilde{v}_{0,1},\tilde{h}_{0,1}\}$,...,$\{\tilde{v}_{0,M},\tilde{h}_{0,M}\}$with the pretrained model. Each sample consists of states of the visible layer and all the hidden layers.\\
\noindent \textbf{for} $t = 0$ to $T$ (number of iterations) \textbf{do}\\
\indent 1. \textbf{Variational Inference:}\\
\indent \textbf{for} each data sample \textbf{v}$_n,n=1$ to $D$ \textbf{do}\\
\indent\indent Perform a bottom-up pass with 
\begin{gather*}
\nu_j^1 = \sigma\Big( \sum_{i=1}^{n_0} 2W_{ij}^1v_i \Big),\\
\nu_k^2 = \sigma\Big( \sum_{j=1}^{n_1} 2W_{jk}^2\nu_j^1\Big),\\
\cdots\\
\nu_p^{N-1} = \sigma\Big( \sum_{l=1}^{n_{N-2}} 2W_{lp}^{N-1}\nu_l^{N-2} \Big),\\
\nu_q^{N} = \sigma\Big( \sum_{p=1}^{n_{N-1}} W_{pq}^{N}\nu_p^{N-1} \Big),
\end{gather*}
\indent\indent where $\{W_{ij}\}$ is the set of pretrained weights.\\
\indent\indent Set $\mathbold{\mu}=\mathbold{\nu}$ and run the mean-field updates with:
\begin{gather*}
\mu_j^1 \leftarrow \sigma\Big(\sum_{i=1}^{n_0} W_{ij}^1v_i + \sum_{k=1}^{n_2}W_{jk}^2\mu_k^2 \Big),\\
\cdots\\
\mu_j^{N-1} \leftarrow \sigma\Big(\sum_{i=1}^{n_{N-2}} W_{ij}^{N-1}\mu_i^{N-2} + \sum_{k=1}^{n_N}W_{jk}^N\mu_k^N \Big),\\
\mu_j^{N} \leftarrow \sigma\Big(\sum_{i=1}^{n_{N-1}} W_{ij}^{N}\mu_i^{N-1} \Big).
\end{gather*}
\indent\indent Set $\mathbold{\mu}_n = \mathbold{\mu}$. \par
\textbf{end for}\\
\\
\indent 2. \textbf{Stochastic Approximation:}\\
\indent \textbf{for} each sample $m=1$ to $M$ \textbf{do}\par
\indent\indent Running one step Gibbs sampling. Get $(\tilde{v}_{t+1,m},\tilde{h}_{t+1,m})$ from $(\tilde{v}_{t,m},\tilde{h}_{t,m})$\\
\textbf{end for}\\
\\
\indent 3. Parameter Update:\\
\indent $W^1_{t+1} = W^1_t + \alpha_t\Big(\frac{1}{D}\sum_{n=1}^D \mathbf{v}_n(\mathbold{\mu}_n^1)^T- \frac{1}{M}\sum_{m=1}^M \tilde{\mathbf{v}}_{t+1,m}(\tilde{\mathbf{h}}_{t+1,m}^1)^T\Big)$\\
\indent $W^2_{t+1} = W^2_t + \alpha_t\Big(\frac{1}{D}\sum_{n=1}^D \mathbold{\mu}_n^1(\mathbold{\mu}_n^2)^T- \frac{1}{M}\sum_{m=1}^M \tilde{\mathbf{h}}_{t+1,m}^1(\tilde{\mathbf{h}}_{t+1,m}^2)^T\Big)$\\
\indent$\cdots$ \\
\indent $W^{n_N}_{t+1} = W^{n_N}_t + \alpha_t\Big(\frac{1}{D}\sum_{n=1}^D \mathbold{\mu}_n^{n_{N-1}}(\mathbold{\mu}_n^{n_N})^T- \frac{1}{M}\sum_{m=1}^M \tilde{\mathbf{h}}_{t+1,m}^{n_{N-1}}(\tilde{\mathbf{h}}_{t+1,m}^{n_N})^T\Big)$\\
\indent Decrease $\alpha_t$.\\
\noindent \textbf{end for}\\
\\
To facilitate computation, $M$ may be chosen as the number of batches.\\
\subsection{Implementation of the Deep Boltzmann Machine}
The implementation of DBM is in the header file dbm.hpp.\\
\\
I. void \textbf{addLayer}(RBMlayer $\&$l)\\
Add a layer to current DBM. This object of class RBMlayer should store information of layer size, weight, bias, etc. It could also be modified after added to DBM.\\
\\
II. void \textbf{setLayer}(std::vector$<$size$\_$t$>$ rbmSize)\\
Object of class DBM could automatically initialize random weights and biases of each layer by inputting a vector of layer sizes.\\
\\
III. void \textbf{train}(dataInBatch $\&$Data, dataInBatch $\&$label, size$\_$t rbmEpoch, LossType l = MSE, ActivationType a = sigmoid$\_$t)\\
This method trains DBM as classifier.\\
\\
IV. void \textbf{fineTuning}(dataInBatch $\&$data, dataInBatch $\&$label)\\
Fine tuning with mean field inference.\\
\\
V. void \textbf{classify}(dataInBatch $\&$data, dataInBatch $\&$label)\\
Classify the data with DBM. Tesing label is used to compute classification error rate.
\subsection{Summary}
In DBM, each layer receives input from all its adjacent layers. Its training is more complicated than other models.

\section{Multimodal Learning Model}
\begin{figure*}[!t]
\centering
\subfigure[Bimodal Deep Belief Network] {\includegraphics[height=1.2in]{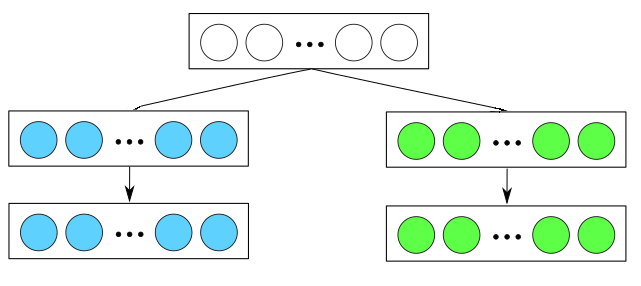}}
\subfigure[Modal Reconstruction Network] {\includegraphics[height=1.8in]{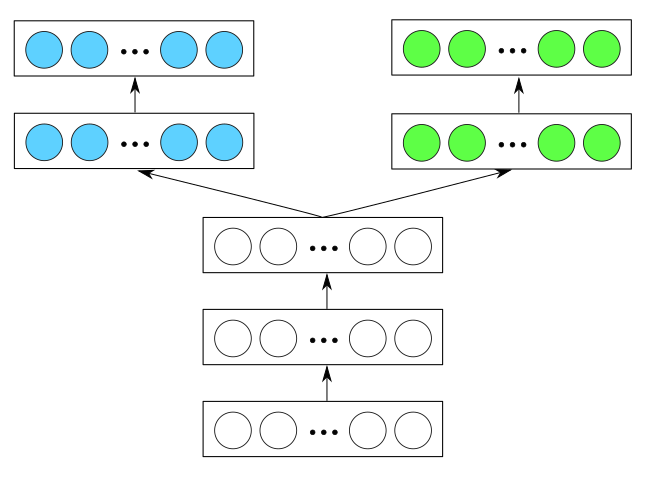}}
\caption{Modal Prediction}
\label{fig5}
\end{figure*}
\begin{figure}[h]
\centering
\includegraphics[height=2.2in]{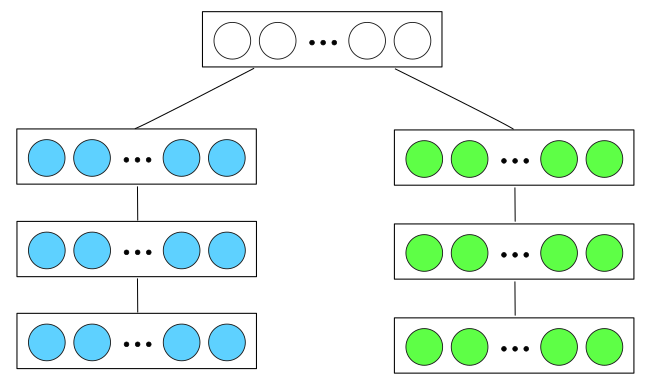}
\caption{MRF Multimodal Learning Model} \label{fig:side:a}
\end{figure}
It is possible to make a low-dimension representation of two modals by building a Bimodal Deep Belief Network \cite{ngiam2011multimodal} as in Figure 8.2(a), in which blue nodes represent data from one modal and green nodes represent data from the other modal. In this process, the recognition weights that are used in bottom-up computation and the generative weights that are used in top-down pass should be learned. If the model is trained with tied weights, half of the memory space could be saved since transposing the weight matrix would transform recognition weights to generative weights. The weights of this model could be used to reconstruct data of two modals as in Figure 8.2(b).\\
\\
Another option is to build a Markov Random Field multimodal learning model \cite{srivastava2012multimodal} by combining two Deep Boltzmann Machines. Figure 8.3 shows such a model. This model is built by first building two DBMs, of which each is trained on data of one modal, then training an RBM on top of these two DBMs.\\
\\
Prediction of data from one modal given data from the other modal could be done by first training a Bimodal Autoencoder in Figure 8.4(a) and then using the modal prediction network in Figure(b) to predict data from two modals. Bimodal Autoencoder should be trained using training data with noises so that it could better predict missing values.
 \begin{figure*}[!t]
\centering
\subfigure[Bimodal Autoencoder] {\includegraphics[height=1.8in]{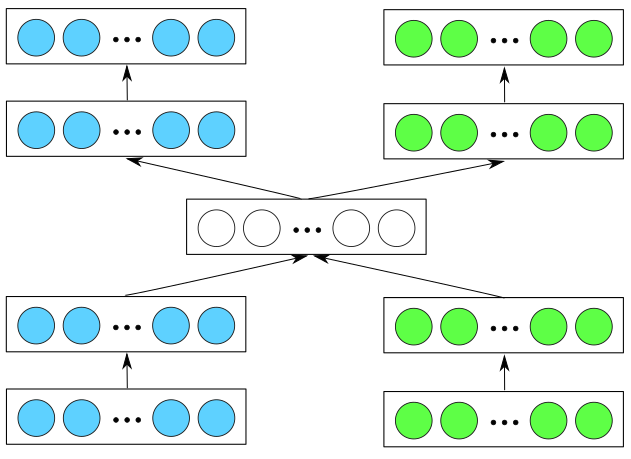}}
\subfigure[Modal Prediction Network] {\includegraphics[height=1.8in]{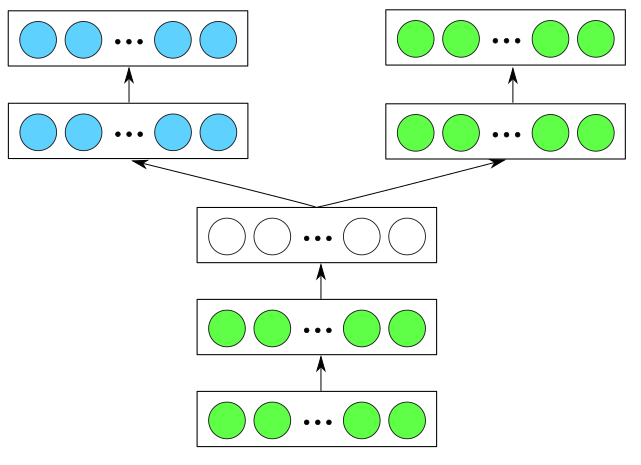}}
\caption{Modal Prediction}
\label{fig5}
\end{figure*}

\section{Library Structure}
\subsection{Data Reading}
\subsubsection{MNIST}
MNIST \cite{lecun1998gradient} is a selected set of samples from NIST data set. It has one training data set, one training label set, one testing data set, and one testing label set. The training set has 60,000 samples and the testing set has 10,000 samples. Each sample data is a 28$\times$28 grey image which is a handwritten integer between 0 and 9. It has 10 classes, so the label is between 0 (inclusive) and 9 (inclusive).\\
\\
The data is stored in big-endian format. The content of data should be read as unsigned characters. Header file readMNIST.hpp provides functions to read this data set.\\
\\
I. cv::Mat \textbf{imread$\_$mnist$\_$image}(const char* path)\\
Read data of MNIST. Each row is a sample.\\
\\
II. cv::Mat \textbf{imread$\_$mnist$\_$label}(const char* path)\\
Read labels of MNIST. Each row is a number indicating the class of that sample.
\subsubsection{CIFAR}
The CIFAR-10 data set \cite{krizhevsky2009learning} consists of 60000 32x32 colour images in 10 classes, with 6000 images per class. There are 5 batches of training images and 1 batch of test images, and each consists of 10000 images.\\
\\
In CIFAR-10 data set, each sample consists of a number indicating its class and the values of the image pixels. The following function in the header file readCIFAR.hpp reads them to four OpenCV matrices:\\
\\
void \textbf{imread$\_$cifar}(Mat $\&$trainingData, Mat $\&$trainingLabel, Mat $\&$testingData, Mat $\&$testingLabel)\\
\\
Each row of the read OpenCV matrices consists of the label and the data of a sample.

\subsubsection{AvLetters}
AvLetters \cite{matthews2002extraction} is the data set recording audio data and video data of different people uttering letters. The dimension of the audio data is 26 and the dimension of the video data is 60$\times$80. The data is stored in single-precision big-endian format. Each file is the data of a person uttering a certain letter. For instance, the file A1$\_$Anya.mfcc contains the audio data of the person named Anya uttering letter "A". \\
\\
The following function in the header file readAvLetters.hpp reads audio data:\\
\\
cv::Mat \textbf{imread$\_$avletters$\_$mfcc}(const char* path)\\
\\
The output is an OpenCV matrix, of which each row contains data of a sample. The original video data is in MATLAB file format. The header file readMat.hpp contains the function\\
\\
cv::Mat \textbf{matRead}(const char* fileName, const char* variableName, const char* saveName).\\
\\
It reads the mat file and at the same time saves it as binary file named as the argument "saveName". This header file uses the MATLAB/c++ interface provided by MATLAB and requires an environment setting, which is contained as comments in the header file. There are some problems running the libraries in this interface together with OpenCV. It would be better to transform all MATLAB files to binary files before training models and then read the transformed binary files. The header file readDat.hpp provides the function to read the transformed binary files:\\
\\
cv::Mat \textbf{readBinary}(const char* file, int rowSize, int colSize)\\
\\
The output is an OpenCV matrix, of which each row contains data of a sample.
\subsubsection{Data Processing}
Header file processData.hpp stores functions processing data.\\
\\
data \textbf{oneOfK}(indexLabel l, int labelDim)\\
Transfer index label to one-of-k expression.\\
\\
dataInBatch \textbf{corruptImage}(dataInBatch input, double denoiseRate)\\
Give corrupted data in batches.\\
\\
std::vector$<$dataInBatch$>$ \textbf{dataProcess}(dataCollection$\&$ reading, int numBatch)\\
Build data batches.\\
\\
dataCollection \textbf{shuffleByRow}(dataCollection$\&$ m)\\
Shuffle the data\\
\\
cv::Mat \textbf{denoiseMask}(size$\_$t rowSize, size$\_$t colSize, double rate)\\
Generate a mask to corrupt data
\subsection{Computation and Utilities}
\textbf{activation.hpp} includes multiple activation functions, such as sigmoid, tanh, relu, leaky$\_$relu, softmax. Each activation function is paired with a function that computes its derivatives to facilitate computation in backpropagation.\\
\\
\textbf{gd.hpp} includes functions for adaptive gradient descent and stochastic gradient descent, as well as a function to anneal the learning rate in which three types of annealing methods are provided.\\
\\
\textbf{inference.hpp} includes the mean field inference implementation used by DBM.\\
\\
\textbf{loss.hpp} includes computation of loss functions. MSE, absolute loss, cross entropy, and binary loss are provided together with the functions to compute their derivatives.\\
\\
\textbf{matrix.hpp} includes some OpenCV matrix manipulation functions.\\
\\
\textbf{loadData.hpp} contains functions to test data loading by visualization.\\
\\
\textbf{visualization.hpp} contains functions of visualization.\\
\subsection{Modules}
Table 1 shows the files that contain modules.
\begin{table}[h]
\centering
\begin{tabular}{|c|c|c|c|c|c|}
\hline
Model & RBM & DNN & DBN & DAE/AE & DBM\\
\hline
Header(.hpp) & rbm &dnn & dbn & autoencoder & dbm\\
\hline
Main(.cpp) & runRBM & runDNN & runDBN & runDAE &runDBM\\
\hline
\end{tabular}
\caption{Files that contain modules}
\end{table}

\section{Performance}
\subsection{Restricted Boltzmann Machine}
Run runRBM.cpp to test the Restricted Boltzmann Machine module. It performs classification on MNIST data set, which has 60,000 training samples and 10,000 testing samples. The hidden layer has 500 units. The classification error rate is 0.0707 (Classification accuracy 92.93$\%$). Multiple deep learning libraries give similar results.
\subsection{Deep Neural Network}
Run runDNN.cpp to test the Deep Neural Network module. It performs classification on MNIST using a DNN with layers of size $(28\times 28)$-500-300-200-10. Without fine tuning, the classification error rate is 0.093 (Classification accuracy 90.7$\%$). With fine tuning, the classification error rate is 0.0288 (Classification accuracy 97.12$\%$). Hinton, Salakhutdinov (2006) claim that DNN with layers of size $(28\times 28)$-500-500-2000-10 could achieve the error rate of 1.2 $\%$. For comparison, a DNN with the same architecture was built and the error rate could achieve $0.0182$. The source codes of the paper use Conjugate Gradient, which is not implemented in MDL. Also, learning is much more sensitive to the learning rate in the $(28\times 28)$-500-500-2000-10 architecture than in the $(28\times 28)$-500-300-200-10 architecture. So finding a better learning rate may also improve the performance.
\subsection{Denoising Autoencoder}
Run runDAE.cpp to test the Denoisinig Autoencoder module. It performs reconstruction of MNIST using a Denoising Autoencoder and an Autoencoder, each with layers of size $(28\times 28)$-500-300-500-$(28\times 28)$. Figure 10.1(a) contains the testing data with noises and the reconstruction given by the Denoising Autoencoder. Figure 10.1(b) contains the testing data without noises and the reconstruction given by the Autoencoder.\\
\\
Fine tuninig reduces more reconstruction error in the Denoising Autoencoder than in the Autoencoder. The reconstruction MSE of the Denoising Autoencoder are: Average error without fine tuning is 6686.69; Average error after fine tuning is 3256.34. The reconstruction MSE of the Autoencoder are: Average error without fine tuning is 4463.24; Average error after fine tuning is 3182.69. With sufficient fine tuning, the reconstruction given by the Denoising Autoencoder is similar to the reconstruction given by the Autoencoder. The visualization is comparable to the published results.\\
\begin{figure*}[!t]
\centering
\subfigure[Denoising Autoencoder] {\includegraphics[height=3.5in]{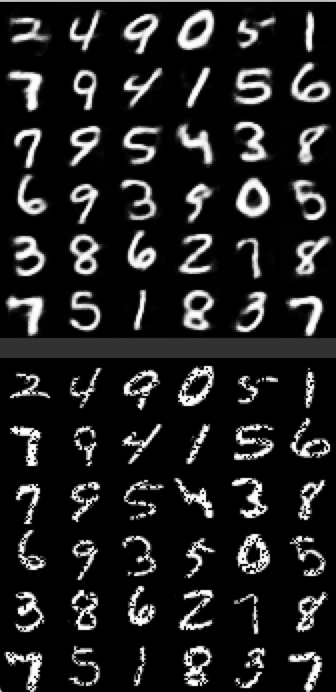}}
\subfigure[Autoencoder] {\includegraphics[height=3.5in]{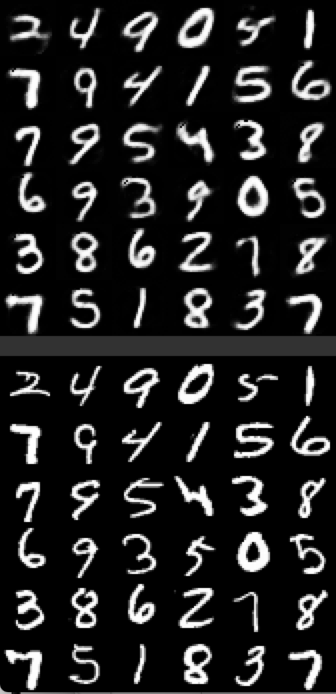}}
\caption{Reconstruction of DAE and AE on MNIST}
\label{fig5}
\end{figure*}

\begin{figure}[h]
\centering
\includegraphics[height=1.3in]{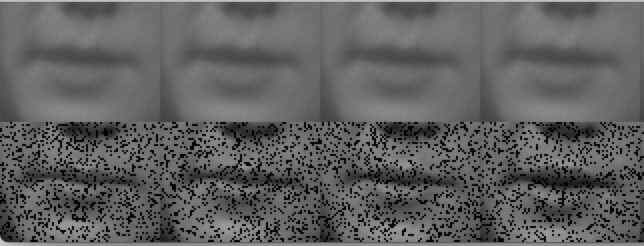}
\caption{Denoising Avletter audio data} \label{fig:side:a}
\end{figure}
Run runDAE$\_$letter.cpp to test the Denoising Autoencoder module with AvLetters data set. Training data consists of the audio data of a person pronouncing the letter ``A" to the letter ``G". Testing data consists of the audio data of the person pronouncing ``H" and ``I". Figure 10.2 shows the result of denosing the testing data.
\subsection{Deep Belief Network}
Run runDBN.cpp to test the Deep Belief Network module. It performs classification on MNIST using DBN with layers of size $(28\times 28)$-500-300-10. The classification error rate on MNIST without fine tuning is 0.0883 (Classification accuracy 91.17$\%$). With the fine tuning using Up-down algorithm, the classification error rate could be reduced to 0.0695 (Classification accuracy 93.05$\%$). Hinton, Osindero $\&$ Teh, (2006) claim that the classification error rate could be as low as 1.25$\%$. Factors such as learning rate decay and gradient descent method could affect the result.
\subsection{Deep Boltzmann Machine}
Run runDBM.cpp to test the Deep Boltzmann Machine module. It performs classification on MNIST using DBM with layers of size $(28\times 28)$-500-500-10. The classification error rate on MNIST without fine tuning is 0.0937 (Classification accuracy 90.63$\%$). Mean Field inference improves the accuracy to 93.47$\%$. Salakhutdinov, Hinton (2009) claim that the classification error rate could be as low as 0.95$\%$. Its source codes use Conjugate Gradient optimization, which is not implemented in this library. This possibly causes the difference.
\subsection{Modal Prediction}
Run modal$\_$prediction.cpp to test the bimodal autoencoder module on AvLetters data set. The training data is the audio and the video data of the letters ``A" to ``G". The audio data of the letters ``H" and ``I" is used to predict the video data of the letters ``H" and ``I". The reconstruction error rate is 10.46$\%$.

\addcontentsline{toc}{section}{References}
\nocite{*}
\bibliographystyle{plain}
\bibliography{mdl}

\end{document}